%% file: main.tex
\documentclass[letterpaper]{article}

\ifdefined\aaaianonymous
    \usepackage[submission]{aaai2026}
\else
    \usepackage{aaai2026}
\fi

\usepackage{times}
\usepackage{helvet}

\usepackage[hyphens]{url}
\usepackage{graphicx}
\usepackage{natbib}
\usepackage{caption}
\usepackage{algorithm}
\usepackage{algorithmic}
\usepackage{amsmath,amssymb}
\usepackage{booktabs}
\usepackage{multirow}
\usepackage{subcaption}
\usepackage{xcolor}
\usepackage{colortbl}
\usepackage{pifont}

\newcommand{\method}{Any-OPD}
\newcommand{\ie}{\textit{i.e.}}

\newcommand{\teasercaption}{\textbf{Any-OPD} enables on-policy distillation between
flow-matching models that share nothing but pixels.
\textbf{Left:} heterogeneous teacher--student pairs (here
FLUX.1-dev~$\to$~SD3.5-Medium) differ in both latent representation
and noise schedule, so no existing on-policy method can supervise
the student's own trajectory.
\textbf{Middle:} \method{} sidesteps both mismatches --- the teacher
refines the student's on-policy sample, and the two models are
compared only in a frozen DINOv2 feature space, never in each
other's latents.
\textbf{Right:} naive latent regression collapses, whereas \method{}
raises the 2.5B student's PickScore from 0.846 to 0.884 and its
HPSv3 from 9.12 to 10.97, approaching or exceeding its 12B teacher
on preference metrics at a fraction of the size.}

\title{\method{}: Heterogeneous On-Policy Distillation for Flow-Matching Models via Representation-Space Bridging}

\author{
    Siming Fu\textsuperscript{\rm 1}\textsuperscript{*},
    Zheming Fu\textsuperscript{\rm 1}\textsuperscript{*},
    Ruizhe He\textsuperscript{\rm 1}\textsuperscript{*},
    Hualiang Wang\textsuperscript{\rm 1},
    Jie Huang\textsuperscript{\rm 1},
    Xiaoxiao Ma\textsuperscript{\rm 1},\\
    Mingchen Zhong\textsuperscript{\rm 1},
    Weihu Huang\textsuperscript{\rm 2},
    Xiaoxuan He\textsuperscript{\rm 2},
    Haojun Xu\textsuperscript{\rm 1}\textsuperscript{\dag}
}
\affiliations{
    \textsuperscript{\rm 1}Joy Future Academy\quad
    \textsuperscript{\rm 2}Zhejiang University \quad
    \textsuperscript{*}Equal contribution.\quad
    \textsuperscript{\dag}Corresponding author.\\
    fusiming.chosen@jd.com, xuhaojun.9@jd.com
}

\makeatletter
\def\@maketitle{%
  \def\theauthors{\if T\showauthors@on\@author\else Anonymous submission\fi}%
  \newcounter{eqfn}\setcounter{eqfn}{0}%
  \vbox{%
    \let\footnote\thanks\relax%
    \setcounter{footnote}{0}%
    \def\equalcontrib{%
      \ifnum\value{eqfn}=0%
        \footnote{These authors contributed equally.}%
        \setcounter{eqfn}{\value{footnote}}%
      \else%
        \footnotemark[\value{eqfn}]%
      \fi%
    }%
    \hsize\textwidth%
    \linewidth\hsize%
    \vskip 0.625in minus 0.125in%
    \centering%
    {\LARGE\bf \@title \par}%
    \vskip 0.1in plus 0.5fil minus 0.05in%
    {\Large{\textbf{\theauthors\ifhmode\\\fi}}}%
    \vskip .2em plus 0.25fil%
    {\normalsize \affiliations_\ifhmode\\\fi}%
    \vskip 0.7em%
    \includegraphics[width=0.98\textwidth]{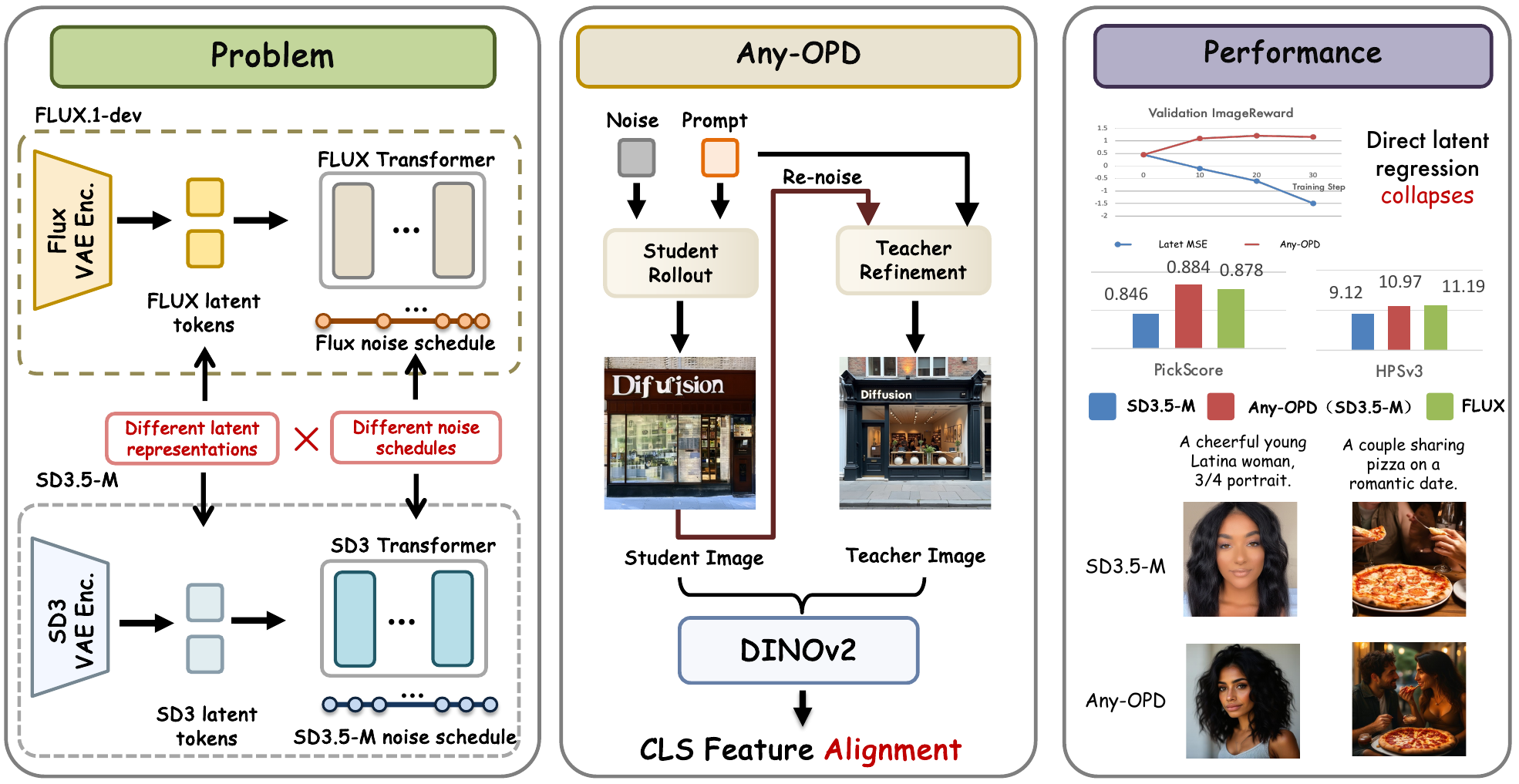}\par%
    \refstepcounter{figure}\label{fig:teaser}%
    \noindent\makebox[\textwidth][c]{%
      \begin{minipage}{0.96\textwidth}%
      \small\textbf{Figure~\thefigure:} \teasercaption%
      \end{minipage}%
    }\par%
    \vskip 1em plus 2fil%
  }%
}%
\makeatother

\nocopyright
\begin{document}
\maketitle

\begin{abstract}
On-policy distillation, in which a teacher corrects samples that the student itself generates, presupposes that the two models speak the same language: identical VAE latents, matching architectures, and a common timestep grid. We ask what happens when none of this holds, as when the strongest teacher available and the student one wishes to deploy come from different model families, and find that the standard recipes have no answer: teacher latents cannot serve as targets in a foreign coordinate system, per-pixel losses against a teacher that stochastically re-draws local detail degenerate into blur or divergence, and timestep indices lose their meaning across mismatched schedules. We present \textbf{\method{}}, to our knowledge the first framework for on-policy distillation between \emph{arbitrary} pairs of latent flow-matching generators. \method{} treats the teacher purely as a black-box sampler and connects the two models at exactly one point: a frozen, model-agnostic vision representation in which their independently decoded outputs are compared, sidestepping every assumption about latents, features, or architecture. Trajectory correspondence is recovered by matching continuous noise levels instead of step indices, and a brief anchoring phase, in which teacher samples are re-encoded through the student's own VAE, ensures the on-policy gradient measures sample quality rather than domain mismatch. Distilling the $12$B FLUX.1-dev into the $2.5$B SD3.5-Medium, \method{} lifts the student's PickScore from $0.846$ to $0.884$ and HPSv3 from $9.12$ to $10.97$, rivaling the teacher at a fifth of its size, where direct latent regression fails to train at all.
\end{abstract}

\input{sec/intro}

\input{sec/related}

\input{sec/method}

\input{sec/experiment}

\input{sec/conclusion}

\bibliography{references}

\end{document}

%% file: sec/intro.tex
\section{Introduction}
\label{sec:intro}

Flow-matching generators~\cite{lipman2023flow,esser2024scaling} now define the state of the art in text-to-image synthesis, but their strongest instances are large: models such as FLUX.1-dev~\cite{flux2024} carry $12$B parameters in the denoising transformer alone, placing them out of reach for latency- or memory-constrained deployment. Knowledge distillation~\cite{hinton2015distilling} offers the natural remedy of letting a small student inherit the quality of a large teacher, and its most effective form is \emph{on-policy}: rather than imitating teacher outputs from fixed data, the student is supervised on states drawn from \emph{its own} sampling trajectory, which eliminates the exposure bias of purely offline imitation~\cite{agarwal2024policy, lin2020autoregressive, ross2011reduction}. Existing diffusion on-policy scheme~\cite{salimans2022progressive, yin2024improved, luo2023latent}, \textbf{however}, rests on an assumption so pervasive that it is rarely stated: the teacher and the student are \textbf{\emph{homogeneous}}. They share a VAE, hence a latent coordinate system in which teacher predictions can serve as regression targets; they typically share an architecture, so internal features can be matched; and they share a timestep grid, so ``the same step'' means the same noise level for both~\cite{salimans2022progressive}. This assumption quietly restricts distillation to pairs derived from a single model family. Yet the pairs one actually wants to distill between are \textbf{heterogeneous}: the available teacher and the student are, in general, developed by different parties around different VAEs, architectures and noise schedules.

In this paper we formulate and address \textbf{heterogeneous on-policy distillation}: train a student on its own trajectories under the supervision of a frozen teacher with which it shares \emph{neither} VAE, \emph{nor} architecture, \emph{nor} noise schedule, accessing the teacher \emph{only as a sampler}. Dropping the homogeneity assumption breaks on-policy distillation along two independent axes. First, the pair has \textbf{no shared space}: a teacher latent is not a valid target in the student's coordinates, and, less obviously, decoding both models to pixels does not restore a usable objective either. Teacher refinement is a stochastic re-synthesis that preserves semantics while re-drawing local detail, so pixel-wise regression pulls toward a blurred conditional mean and, in our experiments, collapses training outright. Second, the pair has \textbf{no shared clock}: the two noise schedules arise from different shift functions, so equal solver indices denote different noise levels~\cite{esser2024scaling}, and index-aligned supervision pairs states from incomparable noise regimes~\cite{karras2022elucidating}.

We introduce \textbf{\method{}} (\emph{any} teacher, \emph{any} student, \emph{o}n-\emph{p}olicy \emph{d}istillation), which resolves each failure with a mechanism that references neither model's internals. For the space gap, the two models are coupled \emph{only} in a third representation external to both: student and teacher each decode with their own VAE, and the resulting images are compared by the CLS embedding of a frozen DINOv2~\cite{oquab2023dinov2}. This global, coordinate-free comparison is invariant to exactly the local re-synthesis that defeats pixel and latent regression, and because the extractor is model-agnostic, \emph{swapping the teacher changes neither the loss nor a single line of the training procedure}. For the clock gap, the two trajectories are aligned by the noise level itself rather than by step index: after the teacher refines the student's sample from a chosen noise level, gradients flow through precisely those student steps that are at least as noisy, a rule that remains well defined for arbitrary pairs of schedules. A third component makes this signal usable from the start: an offline \emph{anchoring} stage first transplants the teacher's output distribution into the student's own latent space via the student's encoder, closing the \emph{distribution-level} gap with data so that the on-policy objective is left to correct only \emph{per-sample} error.
We instantiate \method{} on a maximally heterogeneous pair, distilling FLUX.1-dev~\cite{flux2024} ($12$B) into SD3.5-Medium~\cite{sd35modelcard} ($2.5$B): different VAEs, different transformer architectures, different noise schedules. The distilled student improves substantially over its initialization on preference-oriented metrics, raising PickScore~\cite{kirstain2023pickscore} from $0.846$ to $0.884$ and HPSv3~\cite{ma2025hpsv3} from $9.12$ to $10.97$, and approaches or exceeds the $12$B teacher at roughly one fifth of its size, while the direct latent-regression alternative collapses. \textbf{Our contributions are:}
\begin{itemize}
\itemsep2pt
\item We formulate \textbf{heterogeneous on-policy distillation} for flow-matching generators and identify the two obstructions that make existing on-policy methods inapplicable: the absence of a shared representation (\emph{no shared space}) and of a comparable timestep grid (\emph{no shared clock}).
\item We propose \textbf{\method{}}, which couples the two models solely through a frozen, model-agnostic representation space and aligns their trajectories by \emph{noise-level matching} rather than solver indices. The procedure therefore applies to any teacher--student pair of latent flow-matching generators.
\item We demonstrate the first on-policy distillation across model families, from FLUX.1-dev into SD3.5-Medium, where the $2.5$B student \textbf{matches or surpasses its $12$B teacher} on several preference metrics. We ablate each component and show in particular that naive latent regression across the VAE boundary collapses.
\end{itemize}

%% file: sec/related.tex
%======================================================================
\section{Related Work}
\label{sec:related}

\paragraph{Diffusion and Flow Matching.}
Diffusion models~\cite{ho2020ddpm,song2021scorebased} generate samples by reversing a gradual noising process, while flow matching~\cite{lipman2023flow,liu2023flow} learns a continuous transport from noise to data and supports efficient deterministic sampling. This formulation has become a common backbone for recent high-resolution text-to-image systems, including SD3~\cite{esser2024sd3}, FLUX~\cite{flux2024}, and Z-Image~\cite{cai2025z}. In practice, these models differ not only in scale, but also in VAE design, transformer architecture, and noise-schedule parameterization. Such differences make cross-family distillation substantially more difficult than distillation between homogeneous checkpoints.

\paragraph{Diffusion Distillation.}
Early distillation methods for diffusion and flow models mainly compress sampling cost. Progressive and consistency-based approaches~\cite{salimans2022progressive,luo2023lcd,song2024improved} train few-step students to match a many-step teacher, while distribution matching methods~\cite{yin2024dmd,yin2024dmd2,zhou2024score} align the student's output distribution with teacher preferences. These methods are typically offline: supervision is collected from fixed data or teacher trajectories, so the teacher does not directly correct the states that the student visits during training. On-policy distillation (OPD) was developed in language-model distillation to reduce this exposure bias by letting the student learn from teacher feedback on student-generated trajectories~\cite{agarwal2024policy,gu2024minillm,ko2025distillm}. Recent diffusion and flow extensions bring this idea to iterative image generation. DiffusionOPD~\cite{li2026diffusionopd} formulates on-policy supervision for diffusion transitions, and Flow-OPD~\cite{fang2026flow} adapts OPD to flow matching with dense trajectory-level teacher supervision. However, these formulations assume that teacher and student can interact on shared states, which naturally fits homogeneous models but breaks down when their VAE latent spaces are incompatible.

\begin{figure*}[t]
\centering
\includegraphics[width=0.98\textwidth]{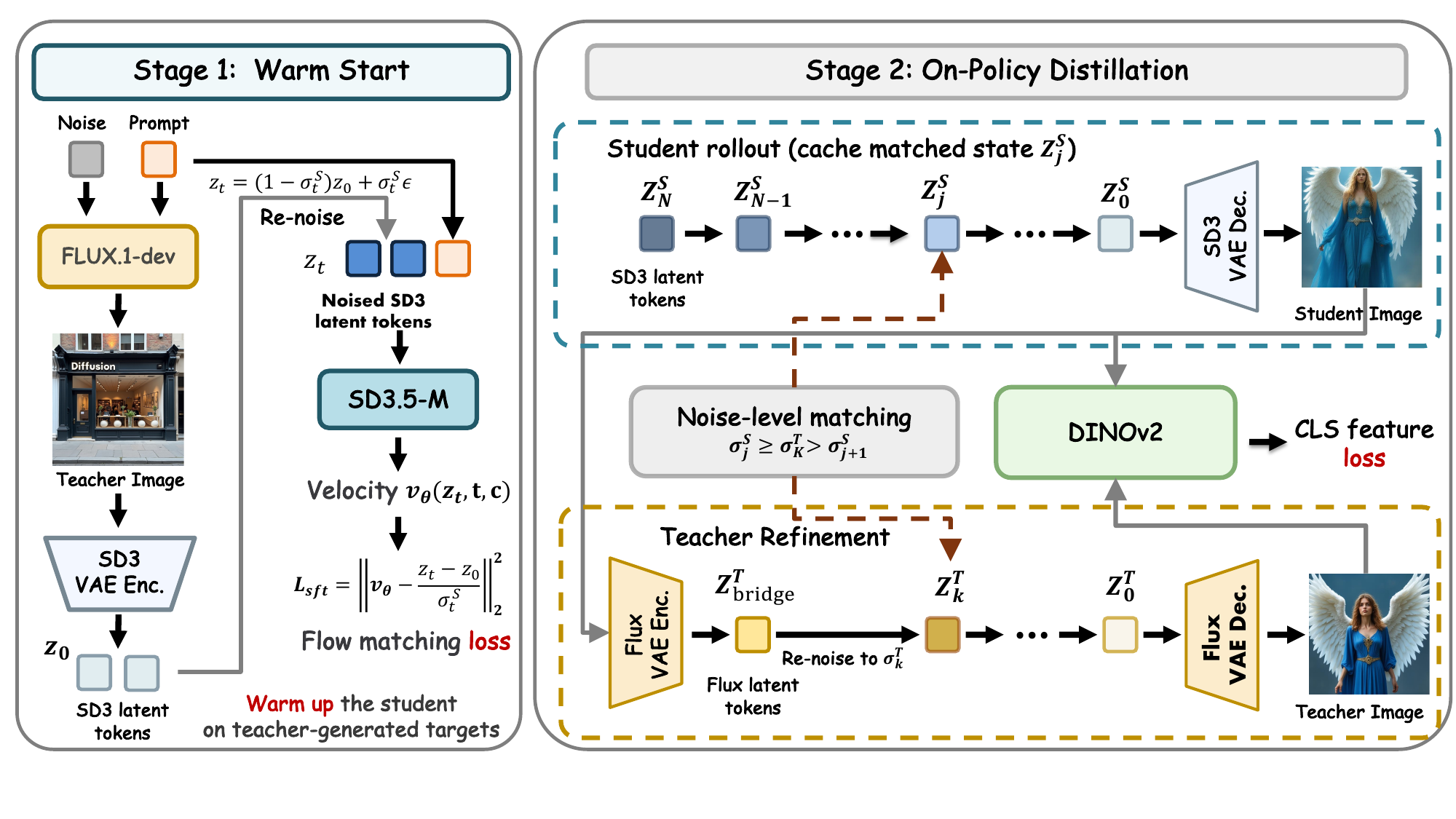}
\vspace{-0.6cm}
\caption{Overview of \method{}. \textbf{Left}: the anchoring stage trains the student by velocity matching on teacher-generated targets, transferring the teacher's distribution into the student's own latent coordinates. \textbf{Right}: on-policy training rolls out the student, projects the sample onto the teacher's manifold by noise-and-denoise refinement, and back-propagates through the student segment selected by noise-level matching. Student and teacher each decode with their own VAE; the two images meet only in a frozen DINOv2 representation space.}
\vspace{-0.5cm}
\label{fig:method}
\end{figure*}

%% file: sec/method.tex
\section{Method}
\label{sec:method}

\subsection{Preliminaries: Flow Matching and Latent Generators}
\label{sec:prelim}

Flow Matching~\cite{lipman2023flow} learns a velocity field $v_\theta:\mathbb{R}^d\times[0,1]\to\mathbb{R}^d$ that transports noise to data along the ODE $\mathrm{d}x_t=v_\theta(x_t,t)\,\mathrm{d}t$. Under the Optimal Transport interpolation,
\begin{equation}
    x_t=(1-t)\,x_0+t\,x_1,\qquad x_0\sim p_{\text{data}},\;\; x_1\sim\mathcal{N}(0,I),
\end{equation}
so that $t$ is itself the noise level, and training minimizes
\begin{equation}
    \mathcal{L}_{\text{FM}}(\theta)=\mathbb{E}_{t,x_0,x_1}\!\left[\left\|v_\theta(x_t,t)-(x_1-x_0)\right\|^2\right].
\end{equation}
Inference uses a discrete $N$-step Euler schedule with noise levels $1=\sigma_0>\sigma_1>\cdots>\sigma_N=0$, where $\sigma_i$ is produced by a model-specific shift function~\cite{karras2022elucidating, esser2024sd3}. Throughout, \emph{solver index $0$ is pure noise and index $N$ is the clean sample}, so $\sigma$ decreases monotonically with the index.

We consider latent generators, each a tuple $(\mathcal{E},\mathcal{D},v,\{\sigma_i\})$ of encoder, decoder, velocity field and schedule: images are produced by integrating $v$ in the latent space $\mathcal{Z}$ and decoding the endpoint~\cite{rombach2022high,esser2024sd3}. Our student is $S=(\mathcal{E}_S,\mathcal{D}_S,v_\theta,\{\sigma^S_i\})$ and our frozen teacher is $T=(\mathcal{E}_T,\mathcal{D}_T,v_\phi,\{\sigma^T_i\})$, with velocity fields acting on disjoint domains,
\begin{equation}
    v_\theta:\mathcal{Z}_S\times[0,1]\to\mathcal{Z}_S,
    \qquad
    v_\phi:\mathcal{Z}_T\times[0,1]\to\mathcal{Z}_T .
\end{equation}

\subsection{The Heterogeneous Setting: No Shared Space, No Shared Clock}
\label{sec:problem}

On-policy distillation requires the teacher to supervise states on the \emph{student's own} trajectory~\cite{ross2011reduction,agarwal2024policy,li2026diffusionopd,fang2026flow}, which for a heterogeneous pair fails along two axes. \textbf{No shared space}: since $\mathcal{Z}_S\not\equiv\mathcal{Z}_T$, a teacher latent is not a valid target in the student's coordinates, and architectural differences rule out feature matching; pixels do not help either, because teacher refinement is a stochastic re-synthesis that preserves semantics while re-drawing local detail, so a pixel-wise MSE regresses toward the mean of all admissible re-syntheses and collapses training~\cite{zhang2018lpips,blau2018perception}. Supervision must move to a \emph{third} representation, external to both models and insensitive to local coordinates. \textbf{No shared clock}: differing shift functions give $\sigma^S_i\neq\sigma^T_i$, so index-aligned supervision pairs states from incomparable noise regimes; the noise level is the only quantity the two trajectories share, and must serve as the alignment variable. These remedies say \emph{how} to supervise, not \emph{when} supervision is useful: while the student's samples remain far from the teacher's manifold, every gradient points out of one domain and into the other, carrying no per-sample information. \method{} therefore \textbf{anchors} the student's distribution offline first, and only then applies the on-policy correction to the per-sample residual.

%----------------------------------------------------------------------
%----------------------------------------------------------------------
\subsection{Anchoring the Student on the Teacher's Manifold}
\label{sec:ws}

The distribution-level gap does not require on-policy training to close, because it does not require correspondence: the student need only learn to produce images of the teacher's kind, not to reproduce any particular one. This makes the gap addressable offline, with the standard flow-matching objective, provided the teacher's distribution can be expressed in the student's coordinates. The student's own encoder provides that expression.

Given a prompt $c$, the teacher generates a target image $x_T=\mathcal{D}_T(\mathrm{Sample}_T(c))$, which we re-encode with the \emph{student's} VAE, $z^{\star}=\mathcal{E}_S(x_T)$. The teacher's manifold is thereby rendered in the coordinate system the student actually optimizes in, something no loss defined across the two latent spaces could achieve. Drawing an index $m$ uniformly from the student's own $M$-step schedule, we form
\begin{equation}
    z_m=(1-\sigma^S_m)\,z^{\star}+\sigma^S_m\,\epsilon,
    \qquad \epsilon\sim\mathcal{N}(0,I),
\label{eq:zt_construct}
\end{equation}
and train with
\begin{equation}
    \mathcal{L}_{\text{WS}}
    =\mathbb{E}_{m,\epsilon,c}\!\left[\left\|v_\theta(z_m,\sigma^S_m,c)-\bigl(\epsilon-z^{\star}\bigr)\right\|^2\right],
\label{eq:ws_loss}
\end{equation}
where $\epsilon-z^{\star}=(z_m-z^{\star})/\sigma^S_m$ by \eqref{eq:zt_construct}; the form above matches $\mathcal{L}_{\text{FM}}$ and avoids dividing by $\sigma^S_m$ as $\sigma^S_m\!\to\!0$. What anchoring cannot fix is equally precise. The student only ever observes states interpolated from a \emph{correct} endpoint, whereas at inference it follows its own imperfect trajectory; anchoring fixes where the student's distribution lies, not how the student behaves on the states it actually reaches. That residual is intrinsically per-sample, and closing it is the task of the on-policy stage.

%----------------------------------------------------------------------
\subsection{On-Policy Distillation between Any Teacher and Any Student}
\label{sec:opd}

The on-policy stage is organized around a single operator. For a noise level $\sigma$, let
\begin{equation}
    \Pi^{\sigma}_T(x)
    \;=\;
    \mathcal{D}_T\!\Bigl(\mathrm{Euler}_T\bigl((1-\sigma)\,\mathcal{E}_T(x)+\sigma\epsilon
    \;\to\;0,\;c\bigr)\Bigr)
\label{eq:projection}
\end{equation}
denote the teacher's noise-and-denoise map: perturb an image to noise level $\sigma$ and regenerate it under the teacher's velocity field~\cite{nie2022diffusion}. Because denoising contracts toward the teacher's model distribution, $\Pi^{\sigma}_T$ acts as a stochastic projection onto the teacher's image manifold~\cite{permenter2023interpreting}, with $\sigma$ setting the radius within which the projection may move the sample: as $\sigma\!\to\!0$ it approaches the identity, and as $\sigma\!\to\!1$ it approaches unconditional resampling~\cite{ho2020ddpm,song2021scorebased}. Evaluating $\Pi^{\sigma}_T$ requires nothing from the teacher beyond the ability to sample; this sampling interface is the entire assumption \method{} places on the teacher, and any latent generator in the sense of Sec.~\ref{sec:prelim} provides it. The stage trains the student so that its own samples become fixed points of this projection up to feature equivalence: a sample $x$ receives zero gradient precisely when $f(x)=f(\Pi^{\sigma}_T(x))$, that is, when the teacher, allowed to redraw the sample within radius $\sigma$, can produce nothing semantically better. Each training iteration accordingly does three things: it projects one on-policy sample, routes the resulting correction through the student steps that can express it, and tests the fixed-point condition. Let $N_S$ and $N_T$ denote the student and teacher solver steps.
\begin{table*}[!t]
\centering
\caption{
\textbf{Main results.}
FLUX.1-dev uses its standard 50-step Euler sampler; SD3.5-Medium and \method{} use 40 Euler steps.
Aesth.: Aesthetic Score; ImgRwd: ImageReward; HPSv3: Human Preference Score v3; UniRwd/UniRwd2: UnifiedReward and UnifiedReward2.
\textbf{Bold} with \colorbox{blue!10}{blue shading} marks improvements over the SD3.5-Medium baseline.
}
\label{tab:main}
\resizebox{\textwidth}{!}{
\begin{tabular}{llc|cccccc|c}
\toprule
& & & \multicolumn{6}{c|}{DrawBench}
& DPG-Bench \\
\cmidrule(lr){4-9}
\cmidrule(lr){10-10}
Role & Model & Params
& Aesth.$\uparrow$
& ImgRwd$\uparrow$
& PickScore$\uparrow$
& HPSv3$\uparrow$
& UniRwd$\uparrow$
& UniRwd2$\uparrow$
& Overall$\uparrow$ \\
\midrule
Teacher & FLUX.1-dev & 12B
& 5.765 & 0.971 & 0.878 & 11.19 & 3.38 & 3.35 & 83.84 \\
Student & SD3.5-Medium & 2.5B
& 5.383 & 0.922 & 0.866 & 9.12 & 3.23 & 3.21 & 84.58 \\
\midrule
Student + \method{} & SD3.5-Medium & 2.5B
& \textbf{\cellcolor{blue!10}5.788}
& \textbf{\cellcolor{blue!10}1.116}
& \textbf{\cellcolor{blue!10}0.884}
& \textbf{\cellcolor{blue!10}10.97}
& \textbf{\cellcolor{blue!10}3.31}
& \textbf{\cellcolor{blue!10}3.26}
& \textbf{\cellcolor{blue!10}84.71} \\
\bottomrule
\end{tabular}
}
\end{table*}

\begin{figure*}[!t]
\centering
\includegraphics[width=\textwidth]{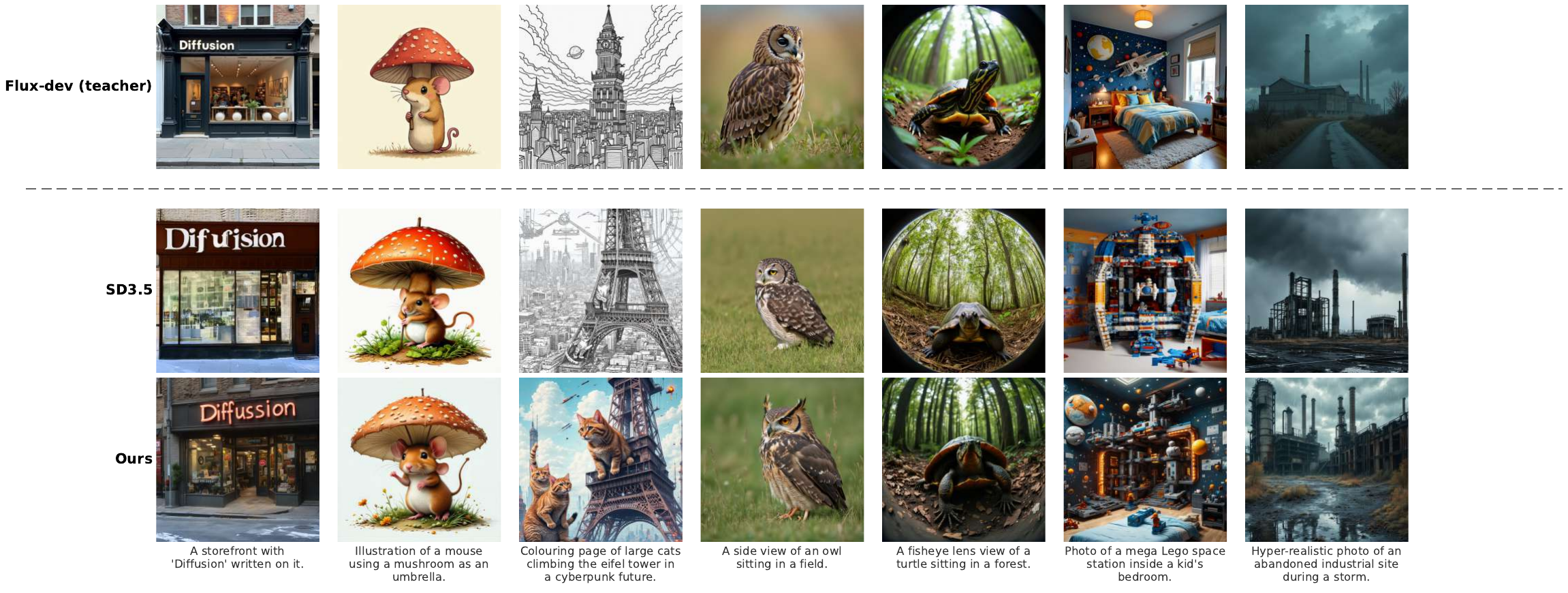}
\caption{Qualitative comparison on DrawBench prompts. \method{} (2.5B, SD3.5-Medium with LoRA) approaches the visual quality of the 12B FLUX.1-dev teacher while keeping the student architecture and 40-step sampling unchanged.}
\label{fig:drawbench_qual}
\end{figure*}

\paragraph{Projecting the on-policy sample.}
Every iteration supervises exactly one sample from the distribution the student deploys at inference. The projection radius is fixed first, by drawing $r\sim\mathcal{U}\{r_{\min},\dots,r_{\max}\}$ and setting the teacher start index $k=N_T-r$; knowing the radius in advance identifies the single intermediate state the later gradient pass will need, so the rollout stores one latent rather than a trajectory. The student then integrates its schedule from $z_0\sim\mathcal{N}(0,I)$ without gradients,
\begin{equation}
    z_{i+1}=z_i+\bigl(\sigma^S_{i+1}-\sigma^S_i\bigr)\,v_\theta(z_i,\sigma^S_i,c),
    \quad i=0,\dots,N_S-1,
\label{eq:euler_rollout}
\end{equation}
and its output is projected at level $\sigma^T_k$. Concretely, $x_S=\mathcal{D}_S(z_{N_S})$ enters the teacher's coordinates through $z^T_{N_T}=\mathcal{E}_T(x_S)$, a format conversion carrying no loss; the teacher re-noises,
\begin{equation}
    z^T_k=(1-\sigma^T_k)\,z^T_{N_T}+\sigma^T_k\,\epsilon,
    \qquad \epsilon\sim\mathcal{N}(0,I),
\label{eq:teacher_noise}
\end{equation}
and integrates its own schedule to the clean endpoint,
\begin{equation}
    x_{\text{ref}}
    =\mathcal{D}_T\!\left(\mathrm{Euler}_T\!\left(z^T_k,\;k\!\to\!N_T,\;v_\phi,\;c\right)\right)
    =\Pi^{\sigma^T_k}_T(x_S).
\label{eq:teacher_refine}
\end{equation}
The radius trades signal against relevance. At small $r$ the projection collapses toward the identity and the target contains no correction; at large $r$ it collapses toward resampling, so $x_{\text{ref}}$ ceases to be an improvement of \emph{this} sample and the gradient inherits the variance of an unpaired target. Useful refinement lives between the two degeneracies, where the projection keeps the student's composition and replaces only its execution; Sec.~\ref{sec:ablation} maps this range empirically.

\paragraph{Routing the correction by noise level.}
A correction has a scale as well as a content. The projection altered the sample only at scales governed by noise levels in $[0,\sigma^T_k]$, so for the student to absorb it, the gradient must reach every solver step of its own that operates in this range; no narrower segment can express the edit. Schedules being incomparable by index, the segment is located by the noise level itself,
\begin{equation}
    j=\max\bigl\{\,i:\;\sigma^S_i\ge\sigma^T_k\,\bigr\},
    \qquad\text{\ie}\quad
    \sigma^S_j\ \ge\ \sigma^T_k\ >\ \sigma^S_{j+1},
\label{eq:sigma_align}
\end{equation}
the least noisy student state still at least as noisy as the projection's start, and hence the tightest segment whose noise range contains the correction's, for arbitrary pairs of shift functions. The student is replayed from the cached $z_j$ with gradients enabled,
\begin{equation}
    \hat{z}_{N_S}=\mathrm{Euler}_S\!\left(z_j,\;j\!\to\!N_S,\;v_\theta,\;c\right),
\label{eq:phase2_rollout}
\end{equation}
so only $N_S-j$ transformer evaluations carry activations, each individually checkpointed. Treating $z_j$ as a constant conditions the gradient on the state the student actually reached, rather than on any state a correct model would have reached; this conditioning, not the rollout alone, is what makes the update on-policy.

\paragraph{Coupling the models in representation space.}
The condition $f(\hat{x}_S)=f(x_{\text{ref}})$ can only be tested in the third representation that Sec.~\ref{sec:problem} identified. Each model renders its result through its own decoder, $\hat{x}_S=\mathcal{D}_S(\hat{z}_{N_S})$ and $x_{\text{ref}}$ from \eqref{eq:teacher_refine}, and the loss couples them through a frozen extractor $f$ applied identically to both:
\begin{equation}
    \mathcal{L}_{\text{OPD}}
    =1-\cos\!\Bigl(f(\hat{x}_S),\;f(x_{\text{ref}})\Bigr),
\label{eq:dino_loss}
\end{equation}
where $f$ returns the CLS token of a frozen DINOv2-Base~\cite{oquab2023dinov2} and gradients flow through the frozen decoder $\mathcal{D}_S$ into $v_\theta$. The choice of $f$ discharges both requirements of Sec.~\ref{sec:problem} at once. It is external: $f$ touches neither model's latents or architecture, so replacing the teacher replaces only the projection $\Pi_T$ and leaves loss and procedure untouched. It is coordinate-free: the CLS token aggregates over all patches, and DINOv2's discriminative self-supervision makes cosine distance penalize structural and semantic deviation while remaining invariant to the local re-synthesis that every application of \eqref{eq:teacher_refine} introduces.

%% file: sec/experiment.tex
\begin{figure*}[t]
\centering
\includegraphics[width=\textwidth]{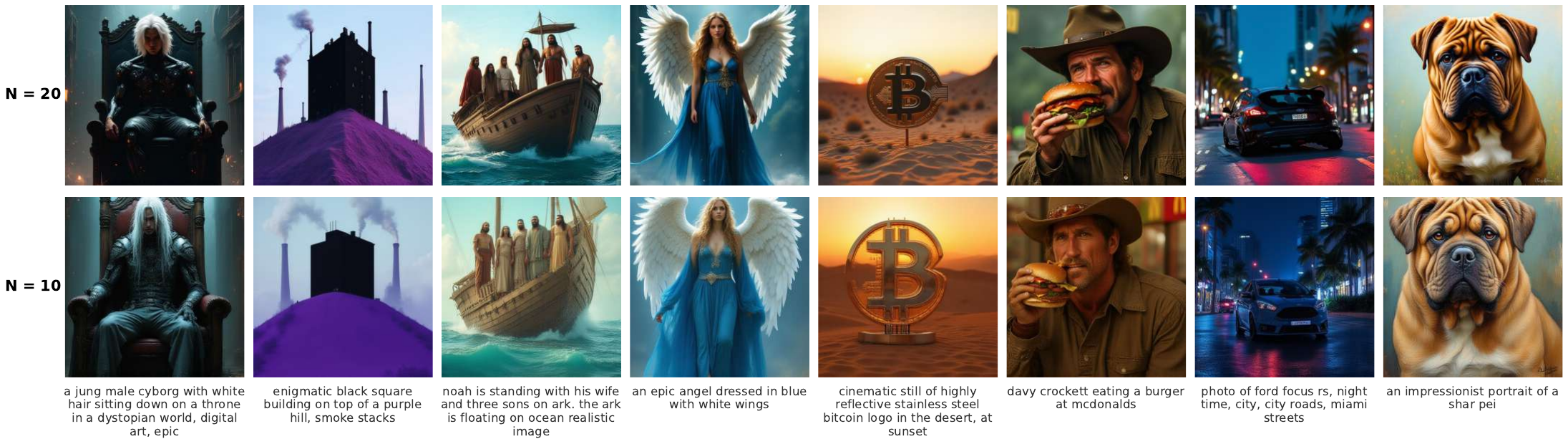}
\caption{Effect of the teacher refinement steps $N_T$ ($10$ vs.\ $20$, otherwise identical). Blur in the $N_T=10$ targets propagates into the student, showing that the teacher's denoising budget bounds the transferable quality.}
\label{fig:denoise_steps}
\end{figure*}

%======================================================================
\section{Experiments}
\label{sec:exp}

\subsection{Experimental Setup}
\label{sec:setup}

\paragraph{Models.}
The student is SD3.5-Medium (2.5B; SD3 VAE, static shift $3.0$, guidance $4.5$); the teacher is FLUX.1-dev (12B; FLUX VAE, dynamic shifting, guidance $3.5$). The pair differs in VAE, architecture and noise schedule simultaneously. We train only LoRA adapters (rank $32$, $\alpha=64$) in the student transformer; everything else stays frozen.

\paragraph{Data and Training.}
Training prompts are sampled from the Pick-a-Pic training set~\cite{kirstain2023pickscore}, with Pick-a-Pic test prompts held out for validation. Anchoring runs for 400 steps ($M=10$ noise levels; teacher targets generated at $512\times512$ with 50-step Euler sampling). On-policy training runs for 800 steps at $512\times512$ with $N_S=N_T=20$. Both phases use learning rate $10^{-4}$ and per-GPU batch size $4$.

\paragraph{Evaluation.}
All models are evaluated at $1024\times1024$ under their standard inference settings (teacher: 50-step Euler; student and \method{}: 40-step Euler). On DrawBench~\cite{saharia2022drawbench} we report Aesthetic Score~\cite{schuhmann2022laion}, ImageReward~\cite{xu2023imagereward}, PickScore~\cite{kirstain2023pickscore}, HPSv3~\cite{ma2025hpsv3}, and UnifiedReward/UnifiedReward2~\cite{wang2025unified} over five images per prompt; on GenEval~\cite{ghosh2023geneval} and \textbf{DPG-Bench}~\cite{hu2024ella} we report the official overall scores over four. Model versions, prompt counts and category-level results are in the supplementary material.

%----------------------------------------------------------------------
\subsection{Main Results}
\label{sec:main_results}
Table~\ref{tab:main} contains the central result: the distilled 2.5B student \emph{overtakes its 12B teacher} on Aesthetic Score, ImageReward and PickScore, and closes most of the gap on HPSv3 and the UnifiedReward scores, improving all six DrawBench metrics over the baseline at an unchanged architecture and sampling budget. That the student can exceed its teacher follows from the supervision itself: the target is the teacher's projection of a \emph{student} sample, combining the student's composition with the teacher's rendering, so training can land the student at points neither model occupies alone; on ImageReward, the gain over the baseline ($+0.194$) is nearly four times the teacher's own margin ($+0.049$). Compositional accuracy is preserved, with DPG-Bench essentially unchanged ($+0.13$): the objective is preference-oriented and the teacher offers no headroom on this benchmark, whereas the compositionally stronger Z-Image teacher does improve it under the same procedure (Sec.~\ref{sec:generalization}). Figure~\ref{fig:drawbench_qual} shows the gains qualitatively: cleaner textures, more coherent lighting and finer structure, with the baseline's layout retained, consistent with a projection that keeps composition and replaces execution.

%----------------------------------------------------------------------
\subsection{Ablation Studies}
\label{sec:ablation}

\subsubsection{The Objective: Only the Representation Space Survives}
\label{sec:ablation_loss}
\begin{figure}[t]
\centering
\begin{subfigure}[t]{0.48\columnwidth}
    \includegraphics[width=\linewidth]{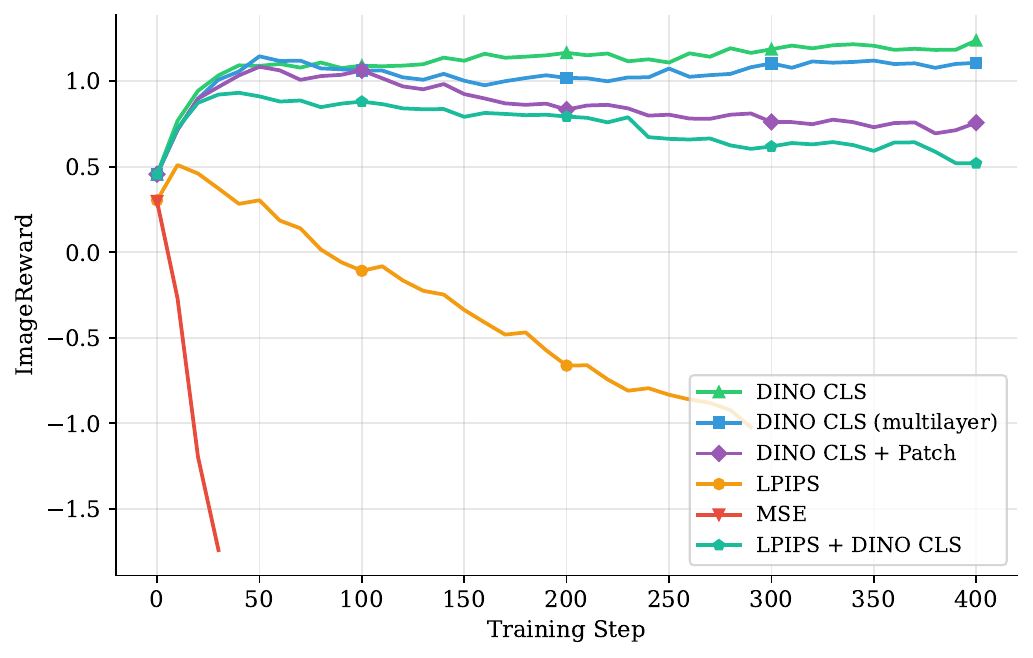}
    \caption{ImageReward}
\end{subfigure}
\hfill
\begin{subfigure}[t]{0.48\columnwidth}
    \includegraphics[width=\linewidth]{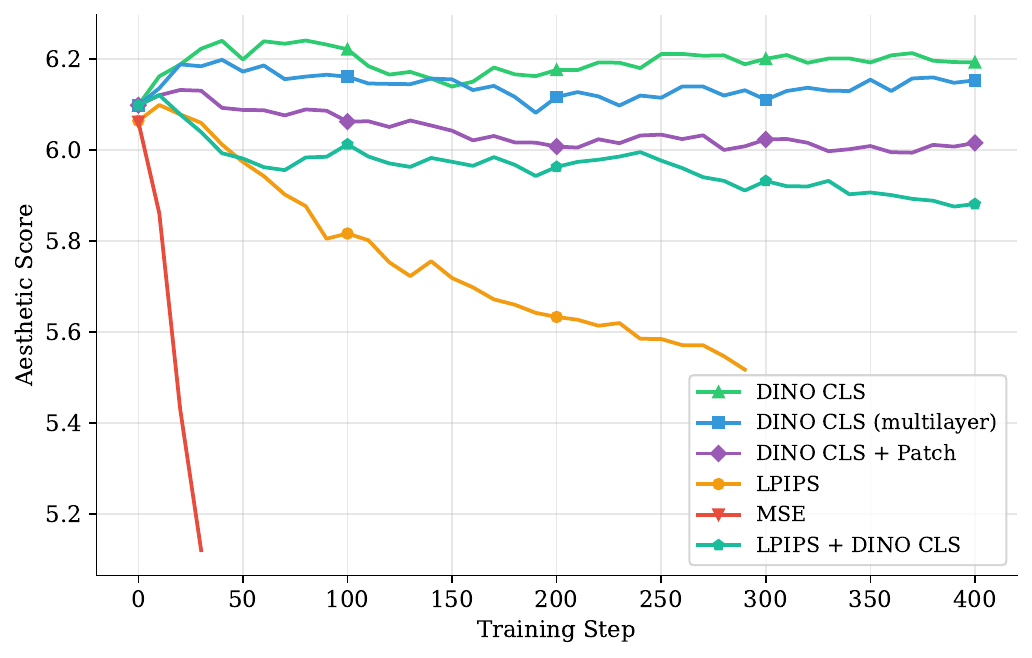}
    \caption{Aesthetic Score}
\end{subfigure}
\caption{OPD training curves under different objectives. Latent MSE collapses; LPIPS degrades late; DINOv2 CLS is stable and best.}
\label{fig:loss_ablation}
\end{figure}

Only the representation-space objective trains stably, confirming the core claim of Sec.~\ref{sec:problem} that heterogeneous pairs can be compared neither coordinate-wise in latents nor locally in pixels. Figure~\ref{fig:loss_ablation} compares OPD trained with latent MSE, LPIPS~\cite{zhang2018lpips}, and three DINOv2 variants (CLS; multi-layer CLS from layers 5, 9, 12; CLS+Patch). Latent MSE collapses within the first training steps on both ImageReward and Aesthetic Score: coordinate-wise regression across the VAE boundary is not merely suboptimal but unstable. LPIPS improves initially, then degrades, since its patch-local features still demand a spatial correspondence that the teacher's stochastic re-synthesis does not respect. All DINOv2 variants remain stable to the end of training, and plain CLS matches or exceeds the richer variants, so a single global embedding suffices. The ordering MSE $<$ LPIPS $<$ DINOv2 tracks exactly how much local correspondence each objective assumes.

\subsubsection{Anchoring Is a Prerequisite, Not a Bonus}
\label{sec:ablation_sft}

\begin{figure}[t]
\centering
\begin{subfigure}[t]{0.48\columnwidth}
    \includegraphics[width=\linewidth]{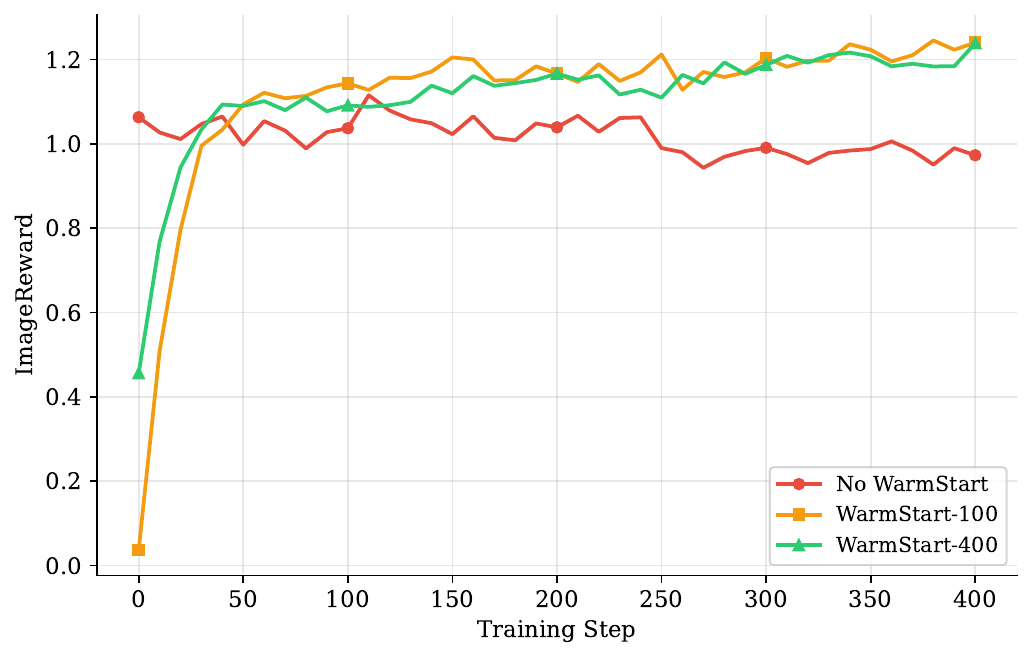}
    \caption{ImageReward}
\end{subfigure}
\hfill
\begin{subfigure}[t]{0.48\columnwidth}
    \includegraphics[width=\linewidth]{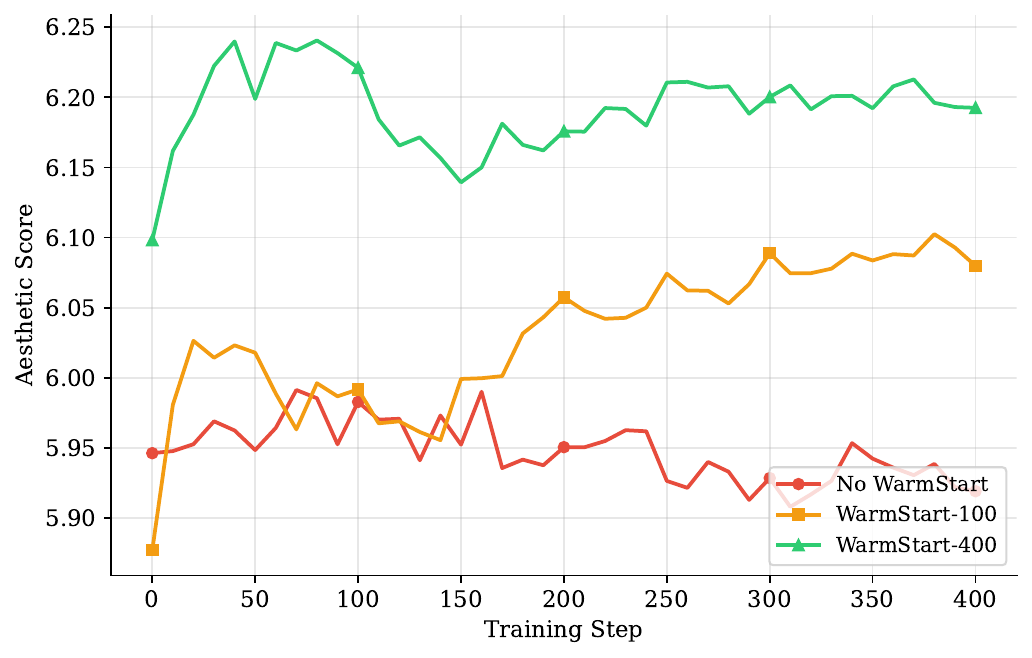}
    \caption{Aesthetic Score}
\end{subfigure}
\caption{OPD curves from different initializations. The stabilized checkpoint (Anchor-400) dominates throughout.}
\label{fig:sft_ablation}
\end{figure}

Anchoring quality strictly determines what OPD can achieve, as Sec.~\ref{sec:ws} predicted. Figure~\ref{fig:sft_ablation} compares OPD started from no anchoring, a pre-stabilization checkpoint (Anchor-100), and a stabilized one (Anchor-400). The three curves are strictly ordered throughout training, and Anchor-100 never catches Anchor-400 despite the identical OPD budget: the gap is one of initialization quality, not total compute, so OPD should begin only after anchoring has converged. Anchoring curves are in the supplementary material.

\begin{figure}[t]
\centering
\begin{subfigure}[t]{0.48\columnwidth}
    \includegraphics[width=\linewidth]{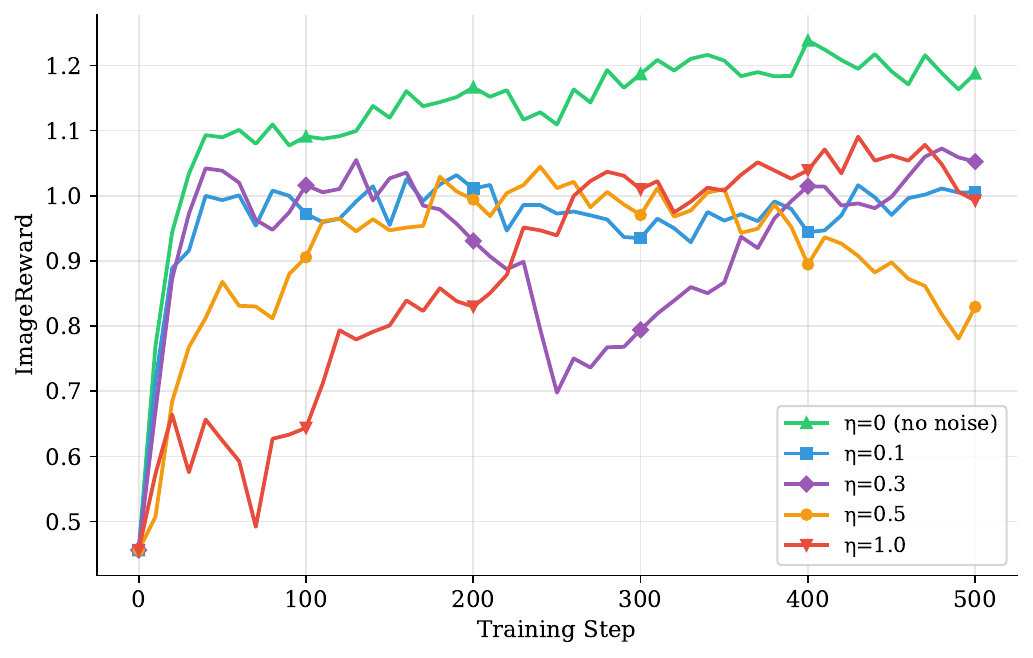}
    \caption{ImageReward}
\end{subfigure}
\hfill
\begin{subfigure}[t]{0.48\columnwidth}
    \includegraphics[width=\linewidth]{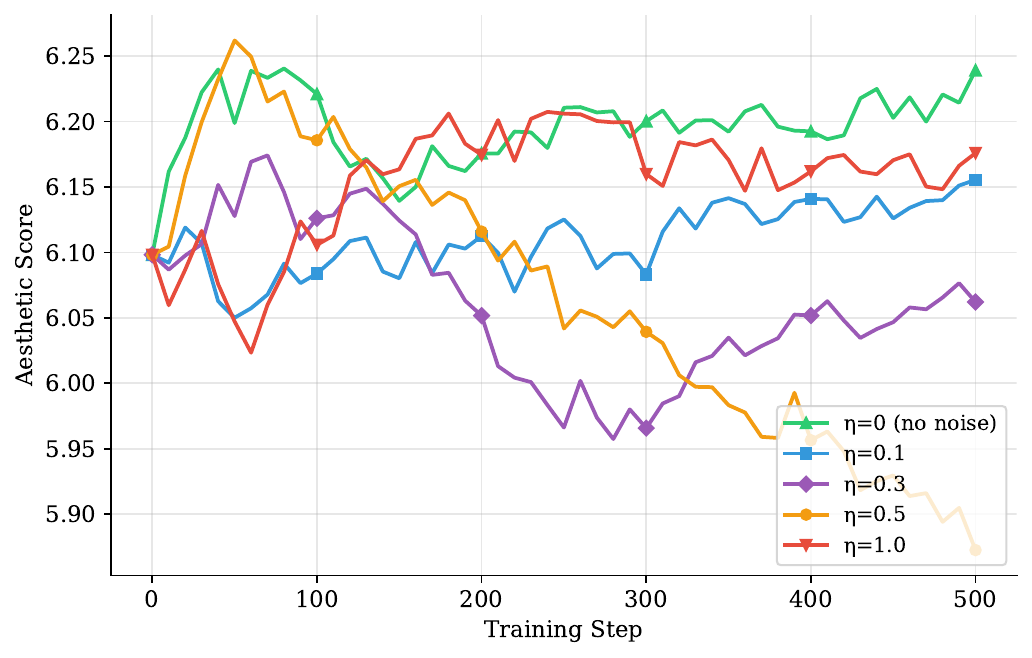}
    \caption{Aesthetic Score}
\end{subfigure}
\caption{OPD curves with CPS rollout stochasticity. Deterministic Euler ($\eta=0$) is best on both metrics.}
\label{fig:cps_ablation}
\end{figure}

\subsubsection{Teacher Refinement Schedule}
\label{sec:ablation_noise}

\paragraph{Refinement strength $r$.}
Strong refinement is necessary: starting the teacher at medium-low noise ($r\in[0,10)$, $\sigma<0.63$) underperforms every other setting on every metric in Table~\ref{tab:ablation_noise}, since the teacher barely alters the sample and the target carries little signal. The two stronger ranges are close, with $[10,15)$ marginally ahead on PickScore and ImageReward and $[15,20]$ winning DPG-Bench and HPSv3; we adopt $[15,20]$ as the best overall balance.

\begin{table}[t]
\centering
\caption{
Effect of the teacher refinement strength on the 20-step teacher schedule. Larger $r$ starts refinement from a noisier level. Stronger refinement yields better results.
}
\label{tab:ablation_noise}
\small
\setlength{\tabcolsep}{4.5pt}
\begin{tabular}{llcccc}
\toprule
& & \multicolumn{3}{c}{DrawBench}
& \multicolumn{1}{c}{DPG} \\
\cmidrule(lr){3-5}
\cmidrule(lr){6-6}
Strength & $r$ Range
& HPSv3$\uparrow$
& Pick$\uparrow$
& ImgRwd$\uparrow$
& Overall$\uparrow$ \\
\midrule
Med-low & $[0,10)$
& 10.24 & 0.881 & 0.891 & 81.21 \\
Med-high & $[10,15)$
& 10.80
& \textbf{\cellcolor{blue!10}0.885}
& \textbf{\cellcolor{blue!10}1.120}
& 84.68 \\
High & $[15,20]$
& \textbf{\cellcolor{blue!10}10.97}
& 0.884 & 1.116
& \textbf{\cellcolor{blue!10}84.71} \\
\bottomrule
\end{tabular}
\end{table}

\paragraph{Refinement steps $N_T$.}
The teacher's denoising budget bounds the transferable quality: in Figure~\ref{fig:denoise_steps}, the halved $N_T=10$ schedule leaves visible blur in the refined targets, which the student faithfully reproduces, while $N_T=20$ yields sharp targets and a correspondingly sharper student.

\subsubsection{Rollout Stochasticity}
\label{sec:ablation_cps}

Rollout stochasticity, though often beneficial in RL-style fine-tuning, hurts here. We replace Euler with Coefficients-Preserving Sampling (CPS)~\cite{wang2025coefficients}, which rotates the noise component of each step from the model-predicted direction $\hat{\epsilon}=x_i+(1-\sigma_i)v_\theta$ toward a fresh Gaussian $\xi\sim\mathcal{N}(0,I)$ by an angle set by $\eta$, at fixed total noise scale $\sigma_{i+1}$:
\begin{equation}
x_{i+1}
=(1-\sigma_{i+1})\,\hat{x}_0
+\sigma_{i+1}\!\left[
\cos\!\tfrac{\eta\pi}{2}\,\hat{\epsilon}
+\sin\!\tfrac{\eta\pi}{2}\,\xi
\right],
\end{equation}
with $\hat{x}_0=x_i-\sigma_i v_\theta$. Deterministic Euler ($\eta=0$) is uniformly best, and $\eta\ge0.3$ degrades after $200$--$300$ steps (Figure~\ref{fig:cps_ablation}): perturbing the rollout decouples the supervised trajectory from the deployed one.

\subsection{Generalization: Swapping the Teacher}
\label{sec:generalization}

The defining claim of \method{} is that the teacher is replaceable without touching the method. We test it literally: FLUX.1-dev is swapped for Z-Image (6B; its own VAE, shift $6$, guidance $5.0$) and \emph{nothing else changes}: same student, loss, hyperparameters and evaluation. Table~\ref{tab:ablation_zimage} shows the student improves on every reported metric, including the compositional benchmarks GenEval ($+0.80$) and DPG-Bench ($+0.95$). Notably, the gains follow the teacher's profile: Z-Image is itself strong on GenEval ($74.90$), and \method{}-Z inherits this strength alongside the perceptual gains.

\begin{table}[t]
\centering
\caption{
Distillation with Z-Image as the teacher, using the identical procedure and hyperparameters as the main experiment.
\textbf{Bold} with \colorbox{blue!10}{blue shading} marks improvements over the SD3.5-Medium baseline; $^\dagger$ marks the teacher reference.
}
\label{tab:ablation_zimage}
\small
\setlength{\tabcolsep}{2pt}
{
\begin{tabular}{lccccc}
\toprule
& \multicolumn{3}{c}{DrawBench}
& \multicolumn{1}{c}{GenEval}
& \multicolumn{1}{c}{DPG} \\
\cmidrule(lr){2-4}
\cmidrule(lr){5-5}
\cmidrule(lr){6-6}
Model
& ImgRwd$\uparrow$
& Pick$\uparrow$
& HPSv3$\uparrow$
& Overall$\uparrow$
& Overall$\uparrow$ \\
\midrule
SD3.5-Medium
& 0.922 & 0.866 & 9.12 & 70.70 & 84.58 \\
Z-Image$^\dagger$
& 0.958 & 0.864 & 9.96 & 74.90 & 86.73 \\
\midrule
\textbf{\method{}-Z}
& \textbf{\cellcolor{blue!10}1.082}
& \textbf{\cellcolor{blue!10}0.872}
& \textbf{\cellcolor{blue!10}9.75}
& \textbf{\cellcolor{blue!10}71.50}
& \textbf{\cellcolor{blue!10}85.53} \\
\bottomrule
\end{tabular}
}
\end{table}

%% file: sec/conclusion.tex
\section{Conclusion}
\label{sec:conclusion}

We introduced \method{}, which extends on-policy distillation beyond the homogeneous setting it has so far been confined to. Our formulation identifies the two assumptions that heterogeneous pairs break, a shared representation and a shared timestep grid, and replaces each with a mechanism that references neither model's internals: the teacher acts purely as a sampler through a noise-and-denoise projection of the student's own outputs, the two models meet only in a frozen external representation where the student is trained toward fixed points of its own projection, and supervision is aligned across incompatible schedules by the noise level itself. A preceding anchoring stage separates distillation by granularity, closing the distribution-level gap with data so that the on-policy objective carries purely per-sample information. Because none of these components depends on the teacher's VAE, architecture or schedule, the teacher is interchangeable by construction, a property we verified by swapping model families without changing a single line of the procedure. We hope this decoupling turns the strongest available generator, whoever builds it, into a usable teacher for whatever model one needs to deploy.

\paragraph{Limitations and Future Work.}
Combining complementary teachers is a natural remedy.  Since nothing in the formulation is specific to images beyond the feature extractor, extending \method{} to other latent generative modalities is a further direction.